\documentclass[10pt,twocolumn,letterpaper]{article}

\usepackage{wacv}
\usepackage{times}
\usepackage{epsfig}
\usepackage{graphicx}
\usepackage{amsmath}
\usepackage{amssymb}
\usepackage{blindtext}
\usepackage{subcaption}

\newcommand\vsa{0mm}
\newcommand\vsg{-2.5mm}
\newcommand\vsp{-0.3cm}

\newcommand*\circled[1]{\raisebox{.7pt}{\textcircled{\raisebox{-1pt} {{#1}}}}}

\usepackage[acronym]{glossaries}
\usepackage{glossaries-extra}
\glsdisablehyper
\newabbreviation{bev}{BEV}{bird's eye view}
\newabbreviation{fps}{FPS}{farthest point sampling}
\newabbreviation[plural=SPGs,firstplural=superpoint graphs (CRFs)]{spg}{SPG}{superpoint graph}
\newabbreviation[plural=CRFs,firstplural=conditional random fields (CRFs)]{crf}{CRF}{conditional random field}
\newabbreviation[plural=CNNs,firstplural=convolutional neural networks (CNNs)]{cnn}{CNN}{convolutional neural network}

\wacvfinalcopy 

\def\wacvPaperID{100} 

\ifwacvfinal\fi
\begin{document}

\title{FuseSeg: LiDAR Point Cloud Segmentation Fusing Multi-Modal Data}

\author{Georg Krispel, Michael Opitz, Georg Waltner, Horst Possegger, Horst Bischof \\
 Institute of Computer Graphics and Vision\\
 Graz University of Technology\\
{\tt\small \{georg.krispel, michael.opitz, waltner, possegger, bischof\}@icg.tugraz.at}
}

\maketitle
\ifwacvfinal\thispagestyle{empty}\fi

\begin{abstract}
We introduce a simple yet effective fusion method of LiDAR and RGB data to segment LiDAR point clouds. Utilizing the dense native \emph{range representation} of a LiDAR sensor and the setup calibration, we establish point correspondences between the two input modalities. Subsequently, we are able to warp and fuse the features from one domain into the other. Therefore, we can jointly exploit information from both data sources within one single network. 
To show the merit of our method, we extend SqueezeSeg, a point cloud segmentation network, with an RGB feature branch and fuse it into the original structure. Our extension called \emph{FuseSeg} leads to an improvement of up to 18\% IoU on the KITTI benchmark. In addition to the improved accuracy, we also achieve real-time performance at 50 fps, five times as fast as the KITTI LiDAR data recording speed.\vspace{-0.3cm}
\end{abstract}

\section{Introduction}
\label{sec:intro}
Being able to segment objects from point clouds is crucial for driver assistant systems, autonomous cars and other robotic perception tasks. Autonomous driving requires multiple sensors to capture all relevant information of the environment. Different types of sensors compensate the individual disadvantages and ensure robust perception in challenging environments. However, fusing and leveraging all this multi-modal data is a non-trivial task.

The task of 3D perception for autonomous vehicles is usually tackled with a combination of RGB cameras and LiDAR sensors (\emph{i.e.} laser range scanners). Recently, numerous architectures with diverse and often complex designs for sensor fusion have been published. However, due to the complexity of this task many methods either use only single-modal input, \emph{e.g.}~\cite{lang2018pointpillars,wu2018squeezeseg,wu2018squeezesegv2} or use the benefits of multi-modalities only after single-modal proposal generation, \emph{e.g.}~\cite{chen2017multi,qi2018frustum,shi2018pointrcnn}. Thus, not all available information is leveraged jointly. Objects poorly visible in one single sensor are prone to be missed. 

To address this problem, we propose a simple and effective fusion method utilizing a dense native representation of laser range scanner data, such that all available information can be processed jointly by common \gls{cnn} architectures. The key idea is to warp expressive RGB features into this LiDAR representation, leveraging correspondences which can be established without any exhaustive search. In this work we focus on the task of point cloud segmentation to show the effectiveness and benefits of our fusion method.

In particular, we extend SqueezeSeg~\cite{wu2018squeezeseg} with an additional branch based on MobileNetV2~\cite{sandler2018mobilenetv2} to leverage RGB information as well. However, na\"ively warping the RGB image into range space and applying an ImageNet \gls{cnn} for early fusion, \emph{e.g.}~\cite{gupta2014learning} or intermediate fusion, \emph{e.g.}~\cite{hazirbas2016fusenet}, hampers the transfer learning benefits of CNNs, as the input image is visually distorted. 

To overcome this issue, we propose to apply the ImageNet \gls{cnn} on the original undistorted RGB image to better leverage the benefits of \glspl{cnn}. Next, we warp the \gls{cnn} features into the range space to get a dense and powerful representation. Thereby, we leverage the RGB/LiDAR calibration to establish \textit{control points} for a polyharmonic spline interpolation~\cite{fasshauer2007meshfree}. We improve SqueezeSeg's segmentation results by a large margin without the use of any synthetic data (in contrast to~\cite{wu2018squeezeseg,wu2018squeezesegv2}). 

We still perform at 50 fps on a NVIDIA GTX 1080Ti GPU, more than twice as fast as common LiDAR sensors dedicated for autonomous cars (typically operating at 20 Hz) and five times as fast as the LiDAR sensor used during the recording of the KITTI benchmark suite (10 Hz). Furthermore, we show that our approach performs better than state-of-the-art RGB semantic segmentation approaches.    

\begin{figure*}
	\centering
	\includegraphics[width=0.9\linewidth]{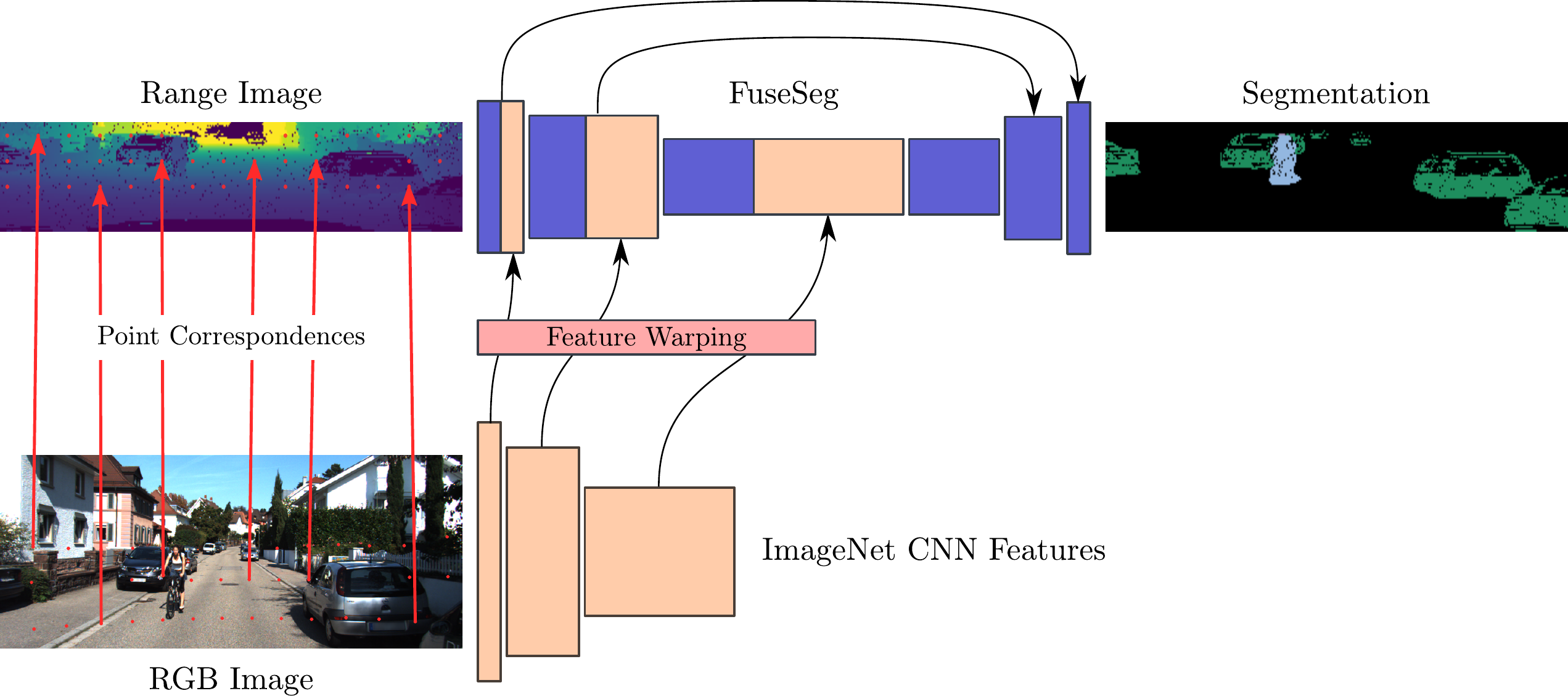}	
	\caption{Schematic overview of our FuseSeg architecture. By exploiting the RGB/LiDAR calibration to establish point correspondences, we fuse feature representations from the RGB and the range image. We utilize the known correspondences to warp the RGB features such that they fit into the range image network. Our range image branch is a slightly modified SqueezeSeg~\cite{wu2018squeezeseg} and we use a MobileNetV2~\cite{sandler2018mobilenetv2} as image branch during our experiments.}\vspace{\vsg}
	\label{fig:fuseseg}
\end{figure*}

\vspace{\vsa}
\section{Related Work}
To better set our work in context, we will first consider recent approaches for 3D point clouds processing (Section~\ref{subsec:full_3d}) and then methods optimized for pseudo-3D/2.5D representations (Section~\ref{subsec:pseudo_3d}). Finally, we will discuss works most related to ours, in particular about the fusion of depth and RGB information (Section~\ref{subsec:rgb3d_fusion}).

\vspace{\vsa}
\subsection{3D Point Cloud Processing}
\label{subsec:full_3d}

Standard \glspl{cnn} require dense input representations on uniform grids. Thus, vanilla \glspl{cnn} can not be used directly on point clouds as they are sparse in 3D space. To overcome this issue, various approaches have been proposed recently. They have been applied to various tasks, \emph{e.g.} classification, 3D object detection and (part-)segmentation. These approaches can be divided into two groups, \emph{i.e.} direct and grid-/graph-based methods. 

\vspace{\vsp}
\paragraph{Direct Methods}\hspace{-0.25cm}are deep architectures which are applied to the point cloud directly. One of the pioneering works in this group is PointNet by Qi~\etal~\cite{qi2017pointnet}. They learn multi-layer perceptrons and linear transformations to map each point individually to an expressive feature space. Subsequently, a max pooling operation generates an order-independent global feature vector, which is utilized for classification and segmentation. 

PointNet lacks the ability to encode local structures with varying density. The subsequent  extension PointNet++~\cite{qi2017pointnet++} tackles this problem by introducing a hierarchical processing strategy. 
Multiple works~\cite{xu2018spidercnn,li2018pointcnn,wang2018deep} introduce a generalization of the classical convolution to irregular point sets. Same as PointNet++, they use a k-nearest neighbor search to overcome the lack of a strictly defined neighborhood.

These methods are able to process a small and fixed amount of points (up to a few thousand). 
To deal with larger point clouds, various strategies like tiling or \gls{fps} must be applied to reduce the amount of processed points. Due to the varying sparsity of LiDAR point clouds, these strategies are usually not very useful when directly applied to single sweeps, as often several samples at nearby salient regions are needed, e.g. to recover an object's outline, instead of few wide-spread samples. For example, the native choice of \gls{fps} are far distant points, which is, given a LiDAR point cloud, not valuable for any downstream task. 

\vspace{\vsp}
\paragraph{Grid-/Graph-based Methods}\hspace{-0.25cm}apply established \glspl{cnn}, transforming the point cloud into grid-based~\cite{riegler2017octnet, maturana2015voxnet, su2018splatnet} or graph-like~\cite{dgcnn, simonovsky2017dynamic} representations. The varying sparsity is the major issue here. Most of the covered space is empty and this would lead to a huge overhead by na\"ively convolving over a regular 3D grid. To enable efficient convolutions, data structures like octrees~\cite{riegler2017octnet}, voxels~\cite{maturana2015voxnet} or high-dimensional lattices~\cite{su2018splatnet} are utilized. These works use sophisticated strategies to avoid redundant computations. However, the required data preprocessing can be time consuming and computationally expensive, especially for larger point clouds.\\

To represent and process large scale point clouds Landrieu and Simonovsky~\cite{landrieu2018large} introduce \glspl{spg}. They transfer the idea of superpixels~\cite{achanta2012slic} to point clouds and propose a geometric pre-partitioning of the data into simple primitives. The resulting \emph{superpoints} are modeled together with derived features within the \gls{spg} and processed with~\cite{simonovsky2017dynamic}.

\vspace{\vsa}
\subsection{Pseudo-3D}
\label{subsec:pseudo_3d}

All considered approaches so far are designed to process sceneries, where objects are fully described in 3D space (\emph{i.e.} both the front and back of an object are reconstructed by the point cloud). However, a single LiDAR sweep just measures depth originating from the sensor center. Thus, it generates a 2.5D representation, where only the surface parts of an object facing the LiDAR are visible. While the point cloud is sparse, in 3D and when projected onto the RGB image plane, a dense representation can be obtained by considering the native properties of the sensor (see Section~\ref{subsec:Preliminaries} for details).

As common LiDARs have a nearly constant horizontal angle resolution, dense representations can be obtained via cylindrical projection~\cite{li2016vehicle,chen2017multi,minemura2018lmnet} or spherical projection~\cite{wu2018squeezeseg,wang2018pointseg,wu2018squeezesegv2}. However, in practice the vertical resolution is not constant. 
For example, the Velodyne HDL-64E laser scanner (used by the KITTI benchmark) sweeps 64 beams with approximately two different angular distances. The top set of 32 beams has a higher angular distance between subsequent beams than the bottom set. Other LiDARs (\emph{e.g.} Velodyne VLP-32C) sample denser near the horizon to improve long-range detections.\\

Our work is based on SqueezeSeg~\cite{wu2018squeezeseg} by Wu~\etal, an adaptation of SqueezeNet~\cite{iandola2016squeezenet} for LiDAR point cloud segmentation. It uses a spherical projection to obtain a dense representation of the LiDAR point cloud and encodes 3D coordinates, range and reflectance intensity into the channels of the input image. In~\cite{wu2018squeezeseg}, they synthesize large amounts of point cloud data utilizing Grand Theft Auto V (GTA-V), a famous video game, to increase its performance on KITTI's \emph{car} class. This synthetic data, however, does not sufficiently represent the other classes realistically, because the underlying geometry has been excessively simplified within the game. For example, the torso, head and limbs of pedestrians within GTA-V are crudely modeled as cylinders. In our work we do not rely on massively generated synthetic data and still achieve state-of-the-art results in real time.
\vspace{\vsa}
\subsection{RGB/3D Fusion}
\label{subsec:rgb3d_fusion}

When depth information is densely available and properly registered with RGB imagery, it is an obvious choice to improve results on different vision tasks. Gupta~\etal~\cite{gupta2014learning} propose three handcrafted auxiliary channels derived from depth to improve segmentation compared to a single depth channel. 
Hazirbas~\etal~\cite{hazirbas2016fusenet} use a separate network branch for depth to improve results compared to an equivalent single branch architectures with additional input channels. Recently, Zeng~\etal~\cite{zeng2019deep} use two network branches to estimate surface normals. 
Similar to these approaches we fuse the respective features at multiple layers as well. However, since depth is not densely available given a LiDAR point cloud, element-wise operations like summation are not sufficient. We introduce a progressive fusion scheme, based on polyharmonic spline interpolation~\cite{fasshauer2007meshfree} to overcome this issue efficiently. 

Recently, various works utilize both RGB and LiDAR data, mostly for the task of 3D object detection. For example, the Multi-View 3D network (MV3D)~\cite{chen2017multi} by Chen \etal maps the LiDAR point cloud to a \gls{bev} to generate object proposals. Given these proposals, features from the \gls{bev}, a cylindrical LiDAR projection and an RGB image branch are fused to classify an object and regress its bounding box.  In Frustum PointNets~\cite{qi2018frustum}, Qi \etal use Faster R-CNN~\cite{ren2015faster} to create 2D proposals from RGB imagery. The result is propagated to 3D space and refined. Except for the object class, there is no further information exchange between the RGB and the 3D detection head. Both works rely on proposals from a single data modality and thus, are prone to loose objects, because they are not using all available information from the beginning on.
Ku~\etal~\cite{ku2018joint} propose \emph{Aggregate View Object Detection (AVOD)}, a network based on RGB and \gls{bev} features. However, they evaluate a predefined set of 3D anchor boxes and thus, are limited by their predefined choice.

Liang~\etal~\cite{liang2018deep} propose a feature warping from an RGB \gls{cnn} branch to a LiDAR \gls{bev}. To this end, they need to perform a k-nearest neighbor search in the point cloud for each pixel in the \gls{bev} image.  However, with the distance to the sensor the point cloud becomes increasingly sparse. In~\cite{liang2019multi} they mitigate this issue utilizing an auxiliary depth completion task. %

However, in contrast to these works, we use two native and dense representations which can be processed by standard \glspl{cnn} without any further preprocessing. Thereby, we are able to densely warp and fuse the features and leverage all information jointly as early as possible.  

\vspace{\vsa}
\section{FuseSeg}

In this section we describe the proposed feature warping module and how we extend SqueezeSeg in order to utilize RGB information. In particular, rather than warping the RGB image into the range space, we apply an ImageNet \gls{cnn} directly on the undistorted input images. Consequently, we can leverage the benefits of transfer learning better, as objects are not distorted in the original RGB input. We then fuse RGB features extracted at multiple layers of the ImageNet \gls{cnn} (MobileNetV2) into the segmentation architecture. 

In order to align the RGB features with the range features for segmentation, we warp them by leveraging the correspondences available due to the calibrated setup. Subsequently, the warped RGB features are concatenated with features from the range image to perform segmentation.

Figure~\ref{fig:fuseseg} schematically illustrates our network architecture and the feature warping. For efficiency, we subsample point correspondences (\textit{control points}) within the different input images. In the following, we discuss the discretization of the LiDAR point cloud (Section~\ref{subsec:Preliminaries}), the foundation of our architecture \textit{SqueezeSeg} (Section~\ref{subsec:squeezeseg}) and the warping procedure (Section~\ref{subsec:feature_fusion}) in more detail.

\vspace{\vsa}
\subsection{LiDAR Geometry}
\label{subsec:Preliminaries}

\begin{figure*}	
	\centering
	\includegraphics[width=0.83\linewidth]{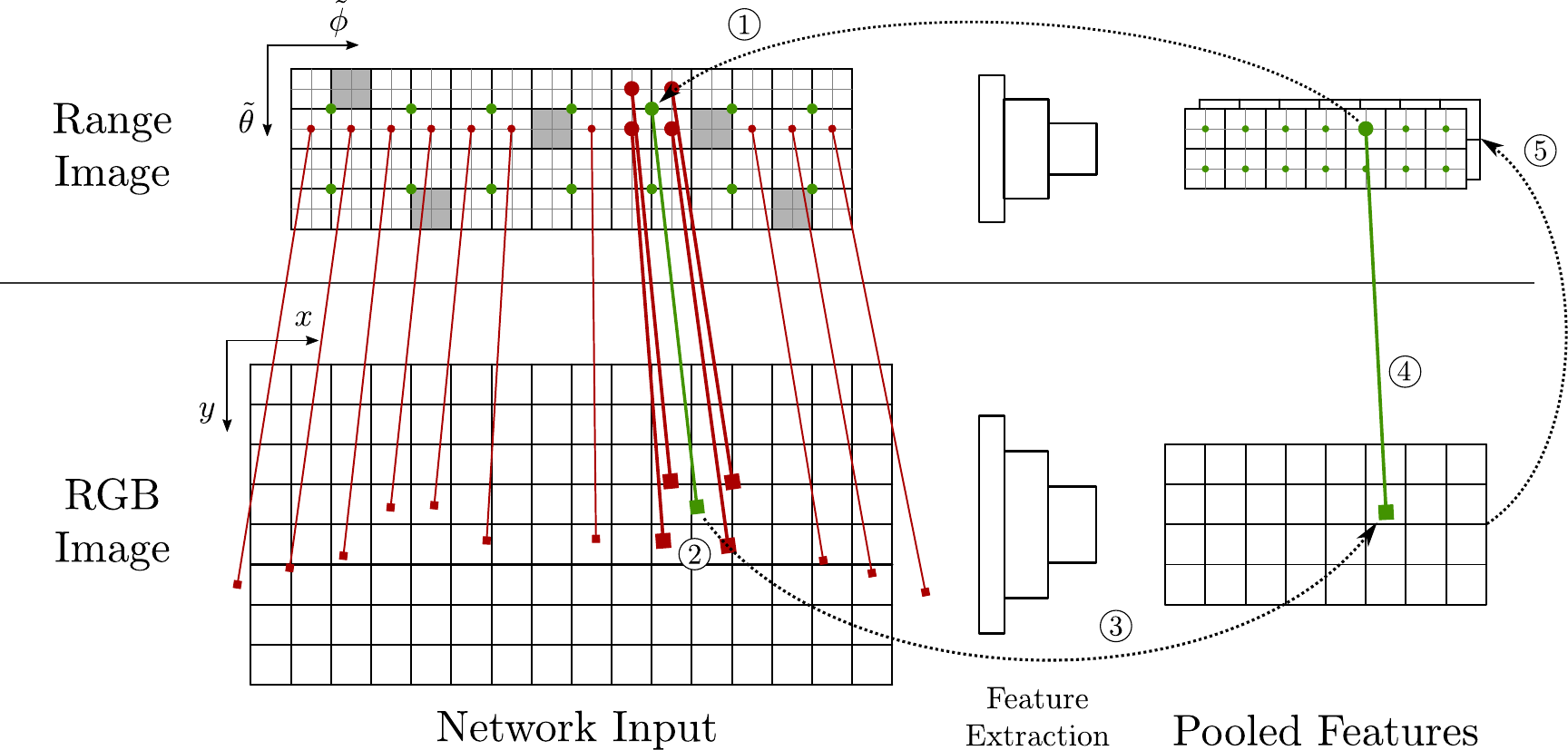}	
	\caption{Illustration of the warping process at a specific feature extraction layer (right). To align the RGB features (bottom) with the range features (top), we first \circled{1}~compute the range image location corresponding to the current range feature (green dots). Given the point correspondences (red) between the range and RGB image, \circled{2}~we use a first-order polyharmonic spline interpolation for sub-pixel sampling of the correct RGB position (green cube). Then, \circled{3} we compute the respective position within the RGB feature space to obtain the feature correspondence \circled{4}. Given that, \circled{5}~we are able to densely warp the RGB features such that they spatially align with the range features. Concatenating them allows for jointly leveraging both information cues for arbitrary 3D perception tasks. The gray pixels denote laser outliers (\emph{e.g.} due to transparent surfaces).}\vspace{\vsg}
	\label{fig:warping}
\end{figure*}

A common LiDAR sensor dedicated for autonomous driving purposes sends out multiple vertically distributed beams and determines the distance to the first hit object by measuring the time-of-flight until the reflection is detected. A $360^{\circ}$ recording is usually obtained by a steady rotation of the laser transmitter itself or a respective deviation \emph{e.g.}, via mirrors. 

SqueezeSeg processes the resulting point cloud on a spherical grid by discretizing the azimuth $\phi$ and zenith $\theta$ of each 3D point $(x, y, z)$ by
\begin{align}
	\phi=&\arcsin \frac{y}{\sqrt{x^{2}+y^{2}}},  \quad\tilde{\phi}=\lfloor\phi / \Delta \phi\rfloor,\\
	\theta=&\arcsin \frac{z}{\sqrt{x^{2}+y^{2}+z^{2}}},\quad\tilde{\theta}=\lfloor\theta / \Delta \theta\rfloor,
\end{align}
where $\Delta \phi$ and $\Delta \theta$ denote the discretization resolution and $(\tilde{\phi}, \tilde{\theta})$ the coordinates on the spherical grid, respectively. The resulting spherical image constitutes a dense representation, which can be processed by a \gls{cnn}. It incorporates five channels, the Cartesian point coordinates~$(x, y, z)$, range $r~=~\sqrt{x^2+ y^2+z^2}$ and the LiDAR's reflectance intensity measurement. Unless stated otherwise, we adopt this channel configuration.

However, in practice the vertical resolution $\Delta \theta$, which is the angle between subsequent LiDAR beams is not constant. Thus, we adapt the representation from~\cite{meyer2019lasernet} and utilize the \emph{beam id} to assign each point to its row $\tilde{\theta}$ in the image. The \textit{beam id} can be easily retrieved from the LiDAR sensor. This allows for an unambiguous vertical discretization to obtain a dense native \emph{range representation}, which we use as the \emph{laser range image}. This range representation is even easier to obtain than the spherical one (\emph{i.e.} no need for zenith projection) and reduces holes and coordinate conflicts in the data. If (due to the horizontal discretization) multiple 3D points fall onto to the same pixel in the range image, we choose the one with azimuth position nearest to the respective pixel center.

\vspace{\vsa}
\subsection{SqueezeSeg}
\label{subsec:squeezeseg}
We base our architecture on SqueezeSeg~\cite{wu2018squeezeseg}. It is a lightweight architecture based on SqueezeNet~\cite{iandola2016squeezenet}, specifically designed to segment spherical images. It adapts the \emph{FireModule} layers from~\cite{iandola2016squeezenet} and introduces related \emph{FireDeconvs} instead of using convolutions and transposed-convolutions in order to reduce computational effort. 

Similar to~\cite{chen2017deeplab}, SqueezeSeg uses a \gls{crf} in order to refine the segmentation results especially at the object borders. The \gls{crf} penalizes assigning different labels to similar points in terms of angular and Cartesian coordinates. In other words, points with nearby coordinates in the range image as well as in 3D space are dedicated to get the same label.

Finally, it minimizes a pixel-wise cross-entropy loss. To mitigate the impact of the class imbalance, cyclists and pedestrians are stronger weighted. Furthermore, outliers, due to failed laser measurements, are masked out during loss computation. 

\vspace{\vsa}
\subsection{Multi-modal Feature Fusion}
\label{subsec:feature_fusion}

In order to merge RGB features from a CNN layer with those from the laser range image, we propose to use the known calibration of LiDAR and RGB camera. We illustrate this process in Figure~\ref{fig:warping}. For each valid pixel in the range image, the corresponding 3D position of the laser point is available. Given the $3\times4$ projection matrix $\mathbf{P}$, we can project the 3D coordinates onto the image via 
\begin{align}
\mathbf{x} &= \mathbf{P}\mathbf{X},
\label{eq:proj}
\end{align}
where $\mathbf{X}$ and $\mathbf{x}$ denote homogeneous 3D and pixel coordinates~\cite{hartley2003multiple}, respectively. The projection matrix itself can be easily derived from the RGB camera calibration and the transformation form LiDAR to camera coordinate system.

Points visible in both, the RGB and range image denote correspondences between the two representations. A na\"ive approach would be to use these correspondences to look up every 3D point's color within the RGB image and thereby colorize the range image. 

However, the comparably dense and valuable information provided by the RGB image would be left unused. Thus, we propose to fuse the intermediate feature representations extracted from respective \glspl{cnn}. We use well studied architectures~\cite{sandler2018mobilenetv2,wu2018squeezeseg} capable of providing useful feature representations for both, the RGB and range image. We extract and warp RGB features at multiple levels of the network such that they align with their range counterparts. We map the ImageNet features from the $7^{th}$, $14^{th}$ and $19^{th}$ layer of MobileNetV2 to the layers \emph{Fire2}, \emph{Fire4} and \emph{Fire7} of SqueezeSeg, respectively. We choose the layers before a pooling operation in MobileNetV2 and warp into similar sized SqueezeSeg layers whilst avoiding the ones which are passed through the skip connections. As a consequence, we exploit the RGB features with the highest representational capabilities of the respective spatial resolution and save parameters within the decoder. Using different or less connection points leads to slightly inferior results.    

Since we warp feature tensors at different network layers (instead of raw input images), we cannot rely on a simple lookup. This is due to the fact that we do not have explicit correspondences between positions within the range feature tensor and their counterparts within the RGB feature tensor. For proper feature warping, we need sub-pixel accuracy (see green line segments in Figure~\ref{fig:warping}). Additionally, we need to deal with laser measurement outliers (e.g. due to transparent surfaces or far distant objects) which cause missing range image-to-RGB correspondences.

To address these issues, we treat the range image-to-RGB correspondences and their positions as \textit{control points} for a first-order polyharmonic spline interpolation~\cite{fasshauer2007meshfree}. Passing query positions $\mathbf{x}$ in the range image, we obtain the corresponding interpolated position $f(\mathbf{x})$ in the RGB image with 
\begin{align}		
	f(\mathbf{x})=\sum_{i=1}^{N} \mathbf{w}_{i} \left|\left|\mathbf{x}-\mathbf{c}_{i}\right|\right|_2+\mathbf{V}^{\mathrm{T}} \left[ \begin{array}{l}{1} \\ {\mathbf{x}}\end{array}\right],
\end{align}
where $\mathbf{c}_{i}$ are the $N$ range pixel coordinates with valid corresponding positions $f(\mathbf{c}_{i})$ in the RGB image. By solving a linear system of equations, we obtain the interpolating spline weights $\mathbf{w}_{i}$ and $\mathbf{V}$. Note that we need to do this computation only once for each sample and we can reuse the weights for all interleaved layers.

In order to retrieve correspondences for a specific spatial resolution, we scale the pixel positions within the range features such that they are aligned with the original input image. Subsequently, we sample the corresponding position in the RGB space using the calculated spline interpolation. This yields the sub-pixel accurate position within the input RGB image for each pixel in the range feature tensor. From this, we can retrieve the corresponding position within the RGB feature tensor as shown in Figure~\ref{fig:warping}. 

To derive the actual value at the non-discrete position in RGB feature space, we bilinearly interpolate the four nearest neighboring features. The part of the warped feature tensor with correspondences outside the RGB image is set to zero.

\vspace{\vsa}
\section{Experiments}
\vspace{\vsa}

\begin{figure*}
  \centering
  \begin{subfigure}[b]{0.48\linewidth}
  	\centering
  	\includegraphics[width=0.95\textwidth]{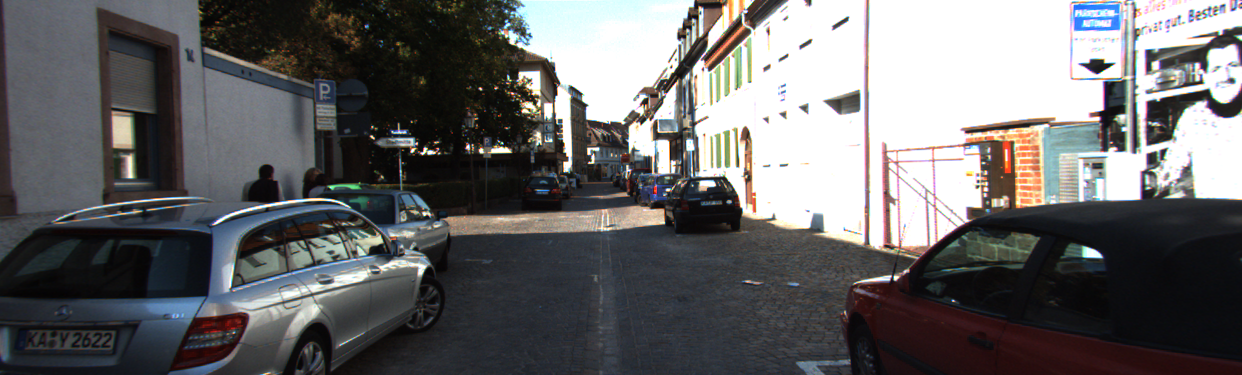}
  \end{subfigure}\hspace{0mm}\begin{subfigure}[b]{0.52\linewidth}
  	\centering
  	\includegraphics[width=\textwidth]{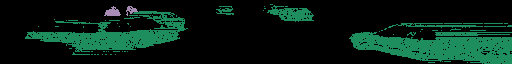}\\[1mm]
  	\includegraphics[width=\textwidth]{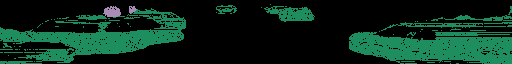}
  \end{subfigure}\\ \small (a) Cars and pedestrians.\\\vspace{0.3cm} 
  \begin{subfigure}[b]{0.48\linewidth}
	\centering
    \includegraphics[width=0.95\textwidth]{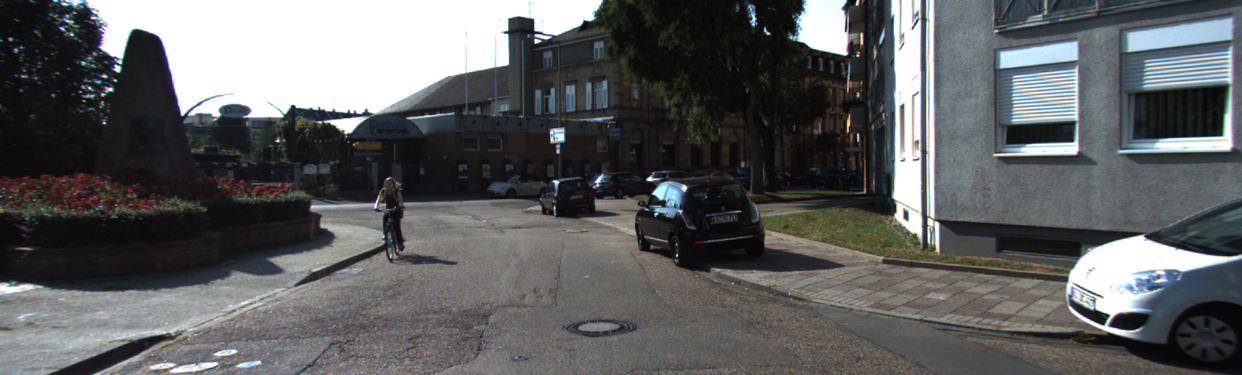}
  \end{subfigure}\hspace{0mm}\begin{subfigure}[b]{0.52\linewidth}
    \centering
    \includegraphics[width=\textwidth]{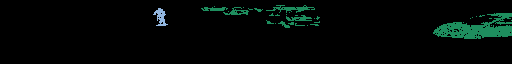}\\[1mm]
    \includegraphics[width=\textwidth]{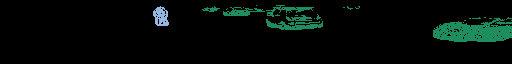}
  \end{subfigure}\\ \small (b) Cars and cyclist.\\\vspace{0.3cm}  
  \begin{subfigure}[b]{0.48\linewidth}
  	\centering
  	\includegraphics[width=0.95\textwidth]{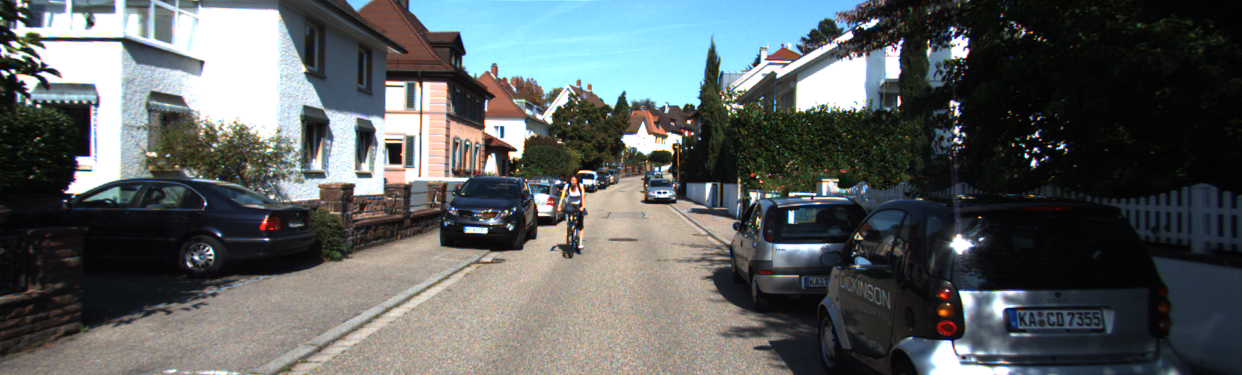}
  \end{subfigure}\hspace{0mm}\begin{subfigure}[b]{0.52\linewidth}
    \centering
    \includegraphics[width=\textwidth]{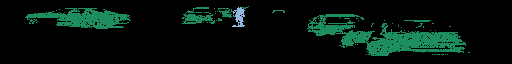}\\[1mm]
    \includegraphics[width=\textwidth]{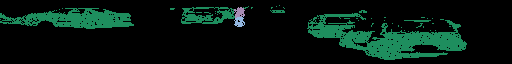}
  \end{subfigure}\\ \small (b) Cars and cyclist.\\\vspace{-0.1cm}  
    	\caption{Qualitative results of FuseSeg. We show the RGB input (left), the ground truth (top right)  and the prediction of the network (bottom right). We detect even small and partially occluded objects~(a,b) as well as objects outside the RGB image and unlabeled in the lower corners of the range image~(a). Sometimes a cyclist is detected separately from the bicycle~(c).}\vspace{\vsg}
    	\label{fig:qualitative}
\end{figure*}

We evaluate our method on KITTI~\cite{geiger2012we, Geiger2013IJRR} and reuse the train/val-split from~\cite{wu2018squeezeseg}. We also follow their training protocol and adopt their parameters: We consider the three main classes \textit{cars}, \textit{pedestrians} and \textit{cyclists} and add an auxiliary class to model the background.
KITTI provides labels in the horizontal field of view of $90^{\circ}$ only, thus we limit our consideration to this area. Additionally, our range images do have the same resolution of $512\times64$ and, unless otherwise stated, the same input channels as in~\cite{wu2018squeezeseg}.

\begin{table}[h]
	\small
	\begin{center}
		\begin{tabular}{|l||c|c|c|c||c|}
			\hline
			Method & car & ped & cyc & avg & rt [ms]\\
			\hline\hline
			FuseSeg & \underline{71.1} & \textbf{36.8} & \textbf{36.0} & \textbf{48.0} & 20 \\	
			FuseSeg R-RGB & 67.4 & 23.4 & 31.2 & 40.7 & 20 \\					
			SqSeg w/o RGB \dag & 67.2 & 20.2 & 24.1 & 37.2 & 9 \\ 
			SqSeg w/ RGB & 63.7 & 18.8 & 22.8 & 35.1 & 13 \\ 
			\hline
			PointSeg~\cite{wang2018pointseg} * & 67.4 & 19.2 & 32.7 & 39.8 & - \\
			SqSeg~\cite{wu2018squeezeseg} * &  64.6 & 21.8 & 25.1 & 37.2 & 13.5 \\
			SqSegV2~\cite{wu2018squeezesegv2} * & \textbf{73.2} & \underline{27.8} & \underline{33.6} & \underline{44.9} & - \\
			\hline
		\end{tabular}
	\end{center}\vspace{\vsg}
	\caption{Point cloud segmentation performance (IoU in \%) and runtime (in milliseconds) on KITTI. To show the effectiveness of our feature fusion we compare with vanilla SqueezeSeg with color as additional input channels. Results marked * are taken from the respective paper and \dag\hspace{0.005cm} mark our reproduced results. Scores and runtime for \emph{SqSeg w/o RGB} differ slightly from~\cite{wu2018squeezeseg} as we retrain it for a fair comparison on our GPU.}\vspace{\vsg}
	\label{tab:eval_squeezeseg}
\end{table}

We augment the data by random horizontal flips and slight deviations in saturation, contrast and brightness of the RGB image. Based on a checkpoint trained with LiDAR features, we re-initialize the respective weights and fine-tune the network. We implement our framework in TensorFlow~\cite{abadi2016tensorflow} and use a GeForce GTX 1080Ti GPU for all runtime evaluations.

In the following, we evaluate the effect of our proposed FuseSeg method on point cloud segmentation in comparison with state-of-the-art methods (Section~\ref{subsec:rgb_features}). Subsequently, we compare the architecture with RGB semantic segmentation networks to validate our warping-based feature fusion (Section~\ref{subsec:sem_seg_comparison}). Finally, we show that we can reduce the number of control points and the accompanied computational cost without negatively affecting the performance (Section~\ref{subsec:num_ctrl_pts}).

\vspace{\vsa}
\subsection{Feature Fusion}
\label{subsec:rgb_features}

\begin{figure*}
	\centering
	\includegraphics[width=0.76\linewidth]{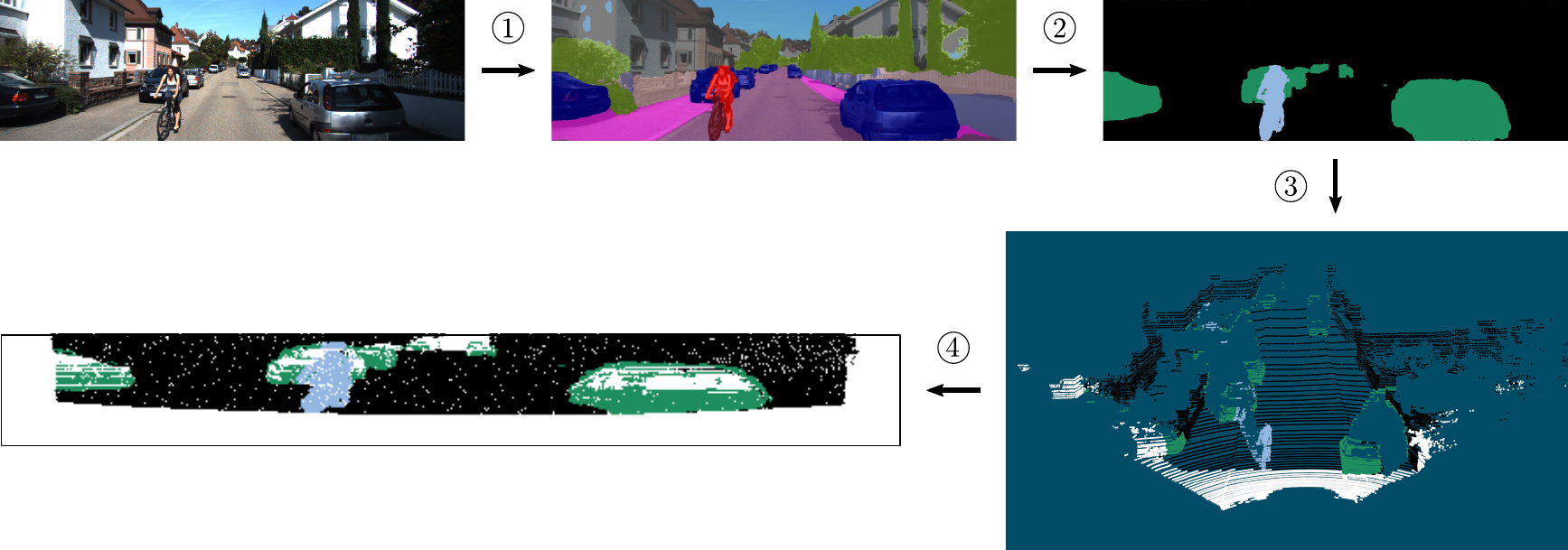}\vspace{-0.1cm}
	\caption{Illustration of the evaluation process solely based on RGB. We infer a semantic mask from the RGB image \circled{1} using a segmentation network trained on KITTI and CityScapes~\cite{Cordts2016Cityscapes}. Subsequently, we fused neighboring \textit{rider} and \textit{bicycle} regions to \textit{cyclist} in order to obtain a compatible annotation policy \circled{2}. Given the calibrated setup and thus, the projection matrix (see Eq.~\ref{eq:proj}) we are able to lookup a class \circled{3} for each point visible in the RGB image (black denotes background, white denotes points with projections outside the RGB image and thus, no derived class). We evaluate the resulting range image \circled{4} (only on the non-white area) and compare it with our method. We show that leveraging depth information using our fusion method significantly outperforms RGB based methods with comparable feature extraction backends, whilst being almost five times faster (see Table~\ref{tab:deeplab}).}\vspace{\vsg}
	\label{fig:rgb_eval}
\end{figure*} 

We show the merit of the fused image features by comparing it not only with SqueezeSeg, but also with state-of-the-art point cloud segmentation methods. Table~\ref{tab:eval_squeezeseg} shows the results for all three relevant object classes and the respective runtime, while Figure~\ref{fig:qualitative} shows some qualitative results. We report the best average intersection-over-union over all three classes.

To provide an additional baseline, we also pass the RGB channels to SqueezeSeg (\emph{SqSeg w/ RGB}). Thus, we colorize its range representation. To this end, we project each point onto the RGB image and sample the underlying pixel's color. Note that not the entire range image is colored, only those 3D points which are visible in the RGB image. 

\begin{figure*}
	\centering
	\includegraphics[width=0.9\linewidth]{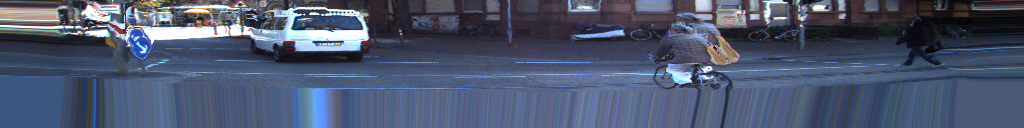}
	\caption{Illustration of warping artifacts due to the baseline between RGB camera and LiDAR sensor. In order to visualize possible artifacts (here \emph{e.g.} cyclist and van roof) we warp the RGB image to its range pendant (we upscale the control points of the range image by a factor of two for visibility). The number and thus, position of control points influence these distortions.}\vspace{\vsg}
	\label{fig:artefacts}
\end{figure*}

The additional color channels even lower the performance of SqueezeSeg. The reason for this drop is that SqueezeSeg is optimized for runtime speed. Consequently its representational power is not able to process all information. Since we utilize a separate lightweight network to process the RGB information, we introduce another baseline (\emph{FuseSeg R-RGB}): We warp the RGB image to its range counterpart (see Figure~\ref{fig:artefacts} for an upscaled example) and pass it to our RGB branch. Note that this baseline has the same number of parameters as FuseSeg.

As we see in our experiments, using a pre-trained ImageNet CNN/MobileNetV2 for extracting features in a warped range image already benefits segmentation performance compared to using no ImageNet CNN for the RGB information. Further, by using our proposed warping method to fuse on the feature level instead of the (RGB) input level, we further significantly improve accuracy. The main reason for this is that the warped RGB input representation is heavily distorted and thus impairs the performance of ImageNet features. In contrast, with our approach the ImageNet CNN operates on an undistorted RGB input on which it better benefits from transfer learning. 

FuseSeg improves segmentation, especially on the smaller classes \emph{pedestrian} and \emph{cyclist}, by a large margin. We increase the mean intersection-over-union (IoU) by 18\% respectively 13.2\% compared to SqueezeSeg. We even outperform its successor SqueezeSegV2~\cite{wu2018squeezesegv2} on average by 3.1\%, which could be improved by our approach as well.

\vspace{\vsa}
\subsection{FuseSeg vs RGB Semantic Segmentation Approaches}
\label{subsec:sem_seg_comparison}

In order to show the effectiveness of our warping-based feature fusion, we compare our approach with semantic segmentation approaches solely relying on RGB information. 
More specifically, we compare FuseSeg with DeepLabv3+~\cite{deeplabv3plus2018} in combination with two feature extraction backends, a MobileNetV2~\cite{sandler2018mobilenetv2} and a more powerful Xception65~\cite{chollet2017xception} feature extractor. We infer that outperforming equivalent state-of-the-art architectures validated our fusion approach. Figure~\ref{fig:rgb_eval} illustrates the process of deriving and evaluating labeled point clouds from RGB segmentation masks.

\begin{table}[h]
	\small
	\begin{center}
		\begin{tabular}{|l||c|c|c|c||c|}
			\hline
			Method & car & ped & cyc & avg & rt [ms]\\
			\hline\hline
			DLv3+ MNV2 & 66.9 &  33.8 & 30.2 & 43.6 & \underline{95}\\
			DLv3+ Xception65 & 71.3 & \textbf{41.4} & \underline{37.4} & \underline{50.0} &369\\
			FuseSeg & \textbf{73.7} & \underline{39.7} & \textbf{41.2} & \textbf{52.1} & \textbf{20}\\
			\hline
		\end{tabular}
	\end{center}\vspace{\vsg}
	\caption{Segmentation performance (IoU in \%) and runtime (in milliseconds) on KITTI. FuseSeg compared with RGB-based semantic segmentation network (DeepLabv3+) trained on both, CityScapes~\cite{Cordts2016Cityscapes} and the KITTI segmentation benchmark. Given the registration, LiDAR points are projected onto the image and classified according to their position in the segmentation mask. We outperform the respective MobileNetV2 (MNV2) DeepLabv3+ by a large margin for all classes and even the much more powerful Xception65 backend on average. Thereby, our architecture is almost five times as fast as the DeepLabv3+ MNV2 counterpart and eighteen times as fast as the Xception65 pendant.}\vspace{-4mm}
	\label{tab:deeplab}
\end{table}

We fine-tune the pre-trained DeepLabv3+ models on CityScapes and the KITTI semantic segmentation data and ensure that no image of our validation set is used for training. We trained until convergence and choose the checkpoint with the best segmentation result on the KITTI validation set. To overcome the diverging annotation policies of the two datasets, we fuse neighboring \emph{bicycle} and \emph{rider} regions to \emph{cyclist}.

We create segmentation masks for each RGB image by passing it through the trained models and segment the 3D points by projecting them onto the masks (see Eq.~\ref{eq:proj}). All classes except \textit{car}, \textit{bicycle} and \textit{pedestrian} are considered as background. Thus, we segment the point clouds without using any depth information. 
For this comparison, we only evaluate the part of the range image with color information for all methods (Thus, the evaluation region differs from Section~\ref{subsec:rgb_features}).

Table~\ref{tab:deeplab} shows the IoU on the respective classes and the runtime of each method. While we clearly outperform DeepLabv3+ in terms of runtime, we outperform the network based on MobileNetV2 on all three classes. Note, that this is the same backend as used in FuseSeg for RGB information. As a consequence, this demonstrates that depth adds valuable information to the segmentation task and our fusion approach is an effective and very efficient method to utilize it.
  
We are even better than the powerful Xception65 DeepLabv3+ on average performance, despite using the weaker backend. Our modular design allows the exchange of the RGB backend in a plug-and-play manner, but one of our research goals is real-time speed leading to the choice of MobileNetV2.

\vspace{\vsa}
\subsection{Number of Control Points}
\vspace{-0.3cm}
\label{subsec:num_ctrl_pts}
\begin{table}[h!]
	\small
	\begin{center}
		\begin{tabular}{|r||c|c|c|c||c|}
			\hline
			\# Ctrl Pts & car & ped & cyc & avg & rt [ms]\\
			\hline\hline
			4\kern 1em  & 69.9 & 33.0 & 33.9 & 45.6 & 19 \\
			24\kern 1em  & 70.4 & 36.2 & \textbf{36.7} & 47.7 & 19 \\
			48\kern 1em  & \textbf{71.1} & \textbf{36.8} & 36.0 & \textbf{48.0} & 20 \\
			96\kern 1em  & 70.7 & 36.0 & 33.8 & 46.8 & 20 \\
			192\kern 1em  & 71.0 & 35.3 & 35.2 & 47.2 & 22 \\
			384\kern 1em  & 70.7 & 36.6 & 36.0 & 47.7 & 26 \\
			
			\hline
		\end{tabular}
	\end{center}\vspace{\vsg}
	\caption{Segmentation Performance (IoU in \%) and runtime (in milliseconds) of FuseSeg on KITTI. We compared different amounts of control points and report best average IoU performance and runtime. While the computational effort of an inference step linearly increases with the amount of control points, performance saturates.}\vspace{\vsg}
	\label{tab:ctrl_pts}
\end{table}

In KITTI there are up to 19k point correspondences between an RGB image and range representation. However, since computational cost of the interpolation increases with the number of control points, a small number of control points is desirable. To this end, to obtain a good coverage in the target domain, we perform \gls{fps} on the coordinates in the range image (in contrast to \gls{fps} on 3D coordinates) to reduce the number of control points.

We compare different configurations aiming at a reliable assessment. We vary the number of control points used by our architecture and evaluate segmentation accuracy as well as runtime. Table~\ref{tab:ctrl_pts} shows the speed-vs-accuracy trade-off. Interestingly, we only need a very small number of control points, \emph{i.e.} 24, to estimate a decent warping and achieve state-of-the-art results. We see that there is no notable variation of the accuracy for the \emph{car} class, which can be explained by their size. 

However, for smaller objects, \emph{i.e.} \emph{pedestrians} and \emph{cyclist}, we observe a notable sensitivity regarding the control points and multiple spikes at certain point numbers. Due to the baseline between camera and LiDAR and the resulting parallax, a flawless warping is not always possible. This distortion peaks at high depth differences, \emph{e.g.} at the edges of visible objects (see Figure~\ref{fig:artefacts}). We hypothesize that a certain number of control points favors these distortions more than others. More elaborate sampling methods, \emph{e.g.} focusing on depth discontinuities within the range
image might mitigate these sensitivities, but are out of the scope of this paper.

\vspace{\vsa}
\section{Conclusion}
We propose a simple and effective way to leverage RGB features for LiDAR point cloud segmentation. Utilizing the \emph{range representation} of LiDAR point clouds allows us to process them with known \gls{cnn} strategies. Then, our efficient warping-based feature fusion enables us to use the benefits of transfer learning on the dense and rich information provided by RGB data jointly with features derived from LiDAR data. Thereby, we still fulfill real-time requirements, performing at 50 fps. This is twice as fast as the recording speed of today's LiDAR sensors. Thus, our method can easily be utilized in autonomous cars and robots. 

Furthermore, the encoder of FuseSeg is applicable as feature extractor for various 3D perception tasks. Finally, our warping strategy in combination with the range representation can be used to interleave features in both directions and thus, also improve RGB-based object detection and semantic segmentation. 

\footnotesize \vspace{-0.4cm}
\paragraph{\footnotesize Acknowledgement}
This project was supported by the Austrian Research Promotion Agency (FFG) project DGT (860820). This work was partially funded by the Christian Doppler Laboratory for Embedded Machine Learning.

{\small
\bibliographystyle{ieee}
\bibliography{egbib}
}

\end{document}